\documentclass[9pt,conference]{IEEEtran}
\IEEEoverridecommandlockouts
\usepackage{cite}
\usepackage{amsmath,amssymb,amsfonts}
\usepackage{algorithmic}
\usepackage{graphicx}
\usepackage{textcomp}
\usepackage{xcolor}
\usepackage{bm}
\usepackage{multirow}
\usepackage{subcaption}
\def\BibTeX{{\rm B\kern-.05em{\sc i\kern-.025em b}\kern-.08em
    T\kern-.1667em\lower.7ex\hbox{E}\kern-.125emX}}
\begin{document}

\title{Goodness-of-pronunciation without phoneme time alignment}

\author{\IEEEauthorblockN{Jeremy H. M. Wong and Nancy F. Chen}
\IEEEauthorblockA{\textit{Institute for Infocomm Research (I\textsuperscript{2}R), A\textsuperscript{$\star$}STAR, Singapore}\\
jeremy\textunderscore wong@i2r.a-star.edu.sg}
}

\maketitle

\begin{abstract}

In speech evaluation, an Automatic Speech Recognition (ASR) model often computes time boundaries and phoneme posteriors for input features. However, limited data for ASR training hinders expansion of speech evaluation to low-resource languages. Open-source weakly-supervised models are capable of ASR over many languages, but they are frame-asynchronous and not phonemic, hindering feature extraction for speech evaluation. This paper proposes to overcome incompatibilities for feature extraction with weakly-supervised models, easing expansion of speech evaluation to low-resource languages. Phoneme posteriors are computed by mapping ASR hypotheses to a phoneme confusion network. Word instead of phoneme-level speaking rate and duration are used. Phoneme and frame-level features are combined using a cross-attention architecture, obviating phoneme time alignment. This performs comparably with standard frame-synchronous features on English speechocean762 and low-resource Tamil datasets.

\end{abstract}


\section{Introduction}

Speech evaluation aims to automatically grade the oral proficiency of a speaker. This is useful in computer-aided language learning scenarios, such as to provide feedback to language learners practising to speak with automatic applications. This paper focuses on evaluation of read speech, where the input is speech and the text that should have been read, and the output is a proficiency score. Like many other tasks, the challenge of data scarcity is often encountered in speech evaluation. This is especially so when developing a system for a low-resource language. In systems such as \cite{huayunzhang2021,gong2022,wong2022,chao2022,wong2023kd,wong2023gpdatauncertainty}, which this paper's setup is an extension of, a frame-synchronous Automatic Speech Recognition (ASR) model is first used to force align the speech with the transcript, to compute time boundaries and phoneme posteriors, which are then used to derive features to perform speech evaluation on. This ASR model needs to be trained on sufficiently diverse paired speech-text data to yield robust performance. The difficulty of collecting such data for low-resource languages hinders the expansion of speech evaluation to these languages. However, the function of the ASR model itself is only for feature extraction, while a separate model is often responsible for the primary speech evaluation task. It thus seems suboptimal that limited availability of ASR training data restricts the development of a speech evaluation system.

The recent developments of speech foundation models offer hope to overcome this limitation. These foundation models are pre-trained on large quantities of diverse data, using either weakly-supervised learning \cite{radford2023} or Self-Supervised Learning (SSL) \cite{baevski2020,hsu2021,chen2022}. This relies on either diversely labelled or unlabelled data, which is easier to obtain than paired speech and annotated transcript data. Furthermore, many speech foundation models are open-source and easily available. These models have been shown to generalise over a broad range of tasks and languages, and may empower the expansion of downstream tasks to low-resource languages.

This paper aims to ease the expansion of speech evaluation to low-resource languages, by overcoming the scarcity of paired speech-text data for ASR model development. The proposal is to leverage existing weakly-supervised multi-lingual speech foundation models to extract the features needed for speech evaluation. However, some multi-lingual weakly-supervised models, such as Whisper \cite{radford2023}, are not frame-synchronous and do not explicitly model phonemes. These hinder the computation of time boundaries and phoneme posteriors. The issues described in more detail are that, first, the model may not express phonemes in its output and may format the text. For example, Whisper outputs byte-level byte-pair encoding \cite{sennrich2016} of the display-formatted text. Second, the frame-asynchronous nature of the model hinders the computation of time alignments. This paper proposes to address these incompatibilities by computing phoneme posteriors from a phoneme-level Confusion Network (CN), obtained from an N-best list generated by the weakly-supervised model. The N-best list is text normalised and mapped to phonemes through a lexicon. It is also proposed to replace the phoneme-level Speaking Rate (SR) and Normalised Duration (ND) features, which require phoneme time alignments, with the word level, for which word time alignments can be computed from the Whisper attention weights.

SSL models offer another way to alleviate data scarcity. Features extracted from these are informative for a variety of downstream tasks. A task-specific downstream model can be trained upon these features on limited labelled data. Using SSL features has been found to often improve the downstream model performance \cite{baevski2020}. For speech evaluation, SSL features have been used in mispronunciation detection \cite{peng2021,wu2021,xu2021b} and predicting other proficiency aspects \cite{banno2023,banno2023a}. The transcription of the speech can also be used as a complementary input to the model \cite{kim2022}. This paper likewise also uses SSL features as one input to a speech evaluation model. In setups such as \cite{huayunzhang2021,gong2022,wong2022,chao2022,wong2023kd,wong2023gpdatauncertainty}, input features to the speech evaluation model are expressed per phoneme. Using per-frame SSL features together with per-phoneme features requires a phoneme time alignment. In \cite{chao2022,wong2023gpdatauncertainty}, the SSL features are first averaged over the frames that are aligned with each canonical phoneme, and then concatenated with the other per-phoneme features. A lack of phoneme time alignments from frame-asynchronous weakly-supervised models hinders such per-phoneme averaging of the SSL features. Rather than averaging the SSL features per phoneme from known alignments, this paper instead proposes to use a cross-attention layer to combine the per-phoneme features with the per-frame SSL features.

These proposals eliminate the need to train a language-specific ASR model for feature extraction, thereby easing the expansion of speech evaluation to new languages that are already supported by multi-lingual open-source models. Such language expansion is demonstrated by experimenting on the low-resource Tamil language. Although the experiments use Whisper in this paper, the proposals are also applicable to other open-source ASR models.

Although these proposals obviate paired speech-text data for ASR model training, speech-text-score data is still needed to train the read-speech evaluation model. The text here can be the transcript that is meant to be read, which is more easily obtainable than an annotation of what the speaker actually uttered. On the other hand, for ASR model training, mismatches between the reference text and speech content may be detrimental to the model performance, and therefore additional effort is required to manually annotate what the speaker actually uttered. For speech evaluation, the reference score is manually annotated here. Works in \cite{liu2023,liang2023,wang2024} consider speech evaluation approaches that do not require annotated scores for training.

\section{Speech evaluation model and features}
\label{sec:features}

\begin{figure*}[t]
\centering
\begin{subfigure}[b]{0.497\textwidth}
\centering
\includegraphics[width=0.55\textwidth]{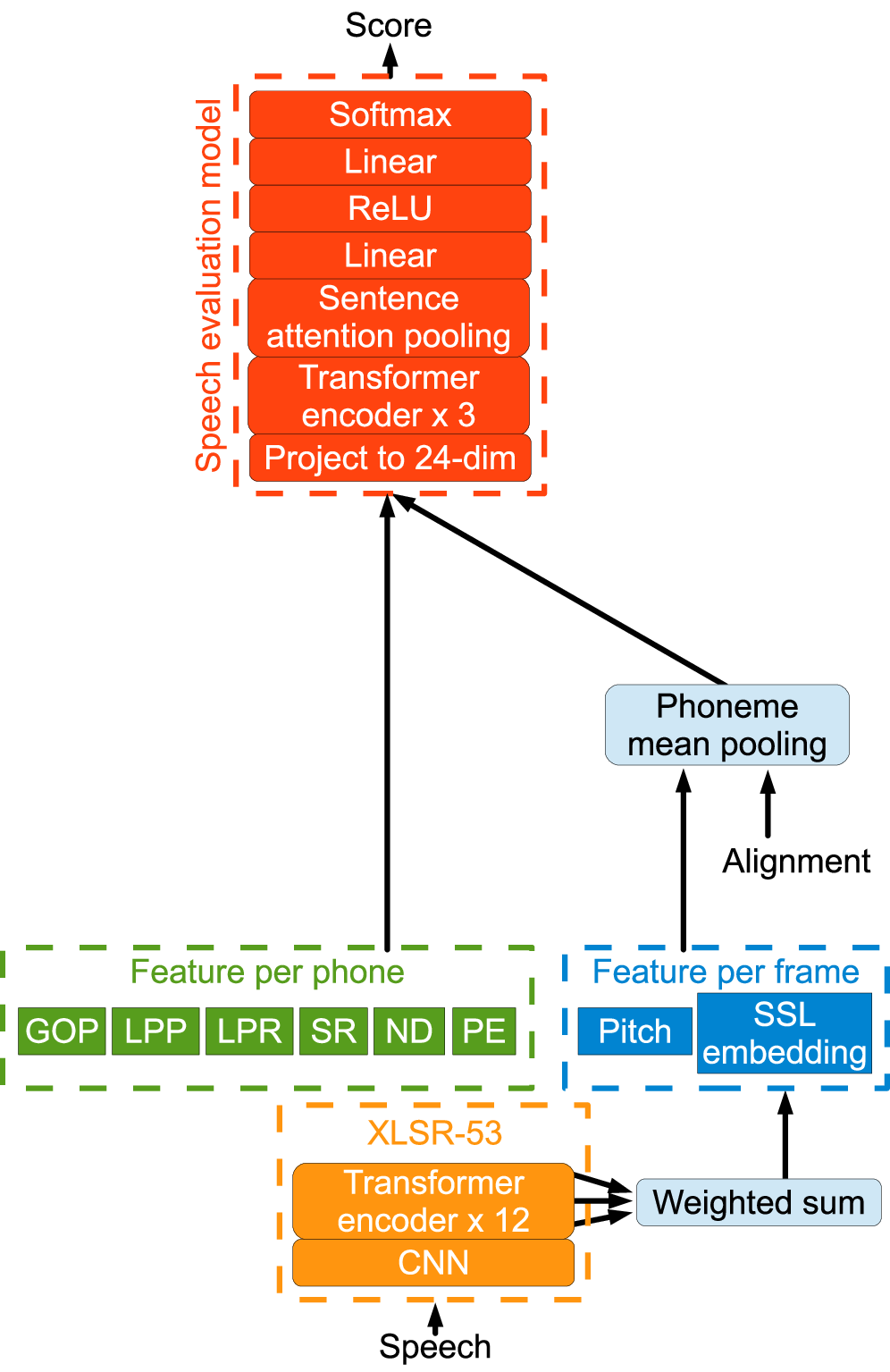}
\caption{Mean pool frame-level features per-phoneme, then concatenate with per-phoneme features}
\label{fig:mean_pool}
\end{subfigure}
\begin{subfigure}[b]{0.497\textwidth}
\centering
\includegraphics[width=0.55\textwidth]{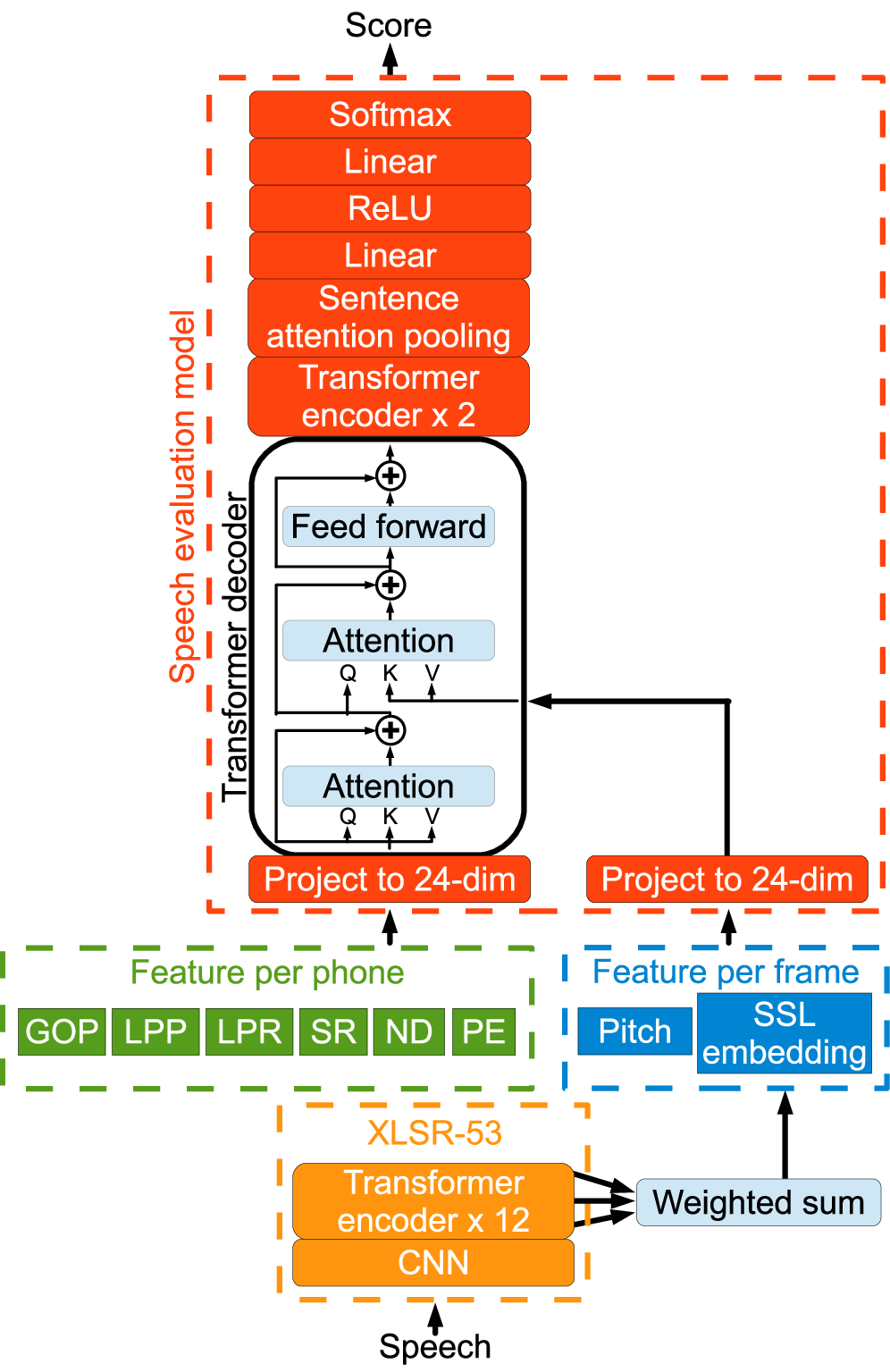}
\caption{Transformer decoder cross-attention to combine per-phoneme and per-frame features}
\label{fig:cross_attention}
\end{subfigure}
\caption{Transformer models for sentence-level speech evaluation}
\label{}
\end{figure*}

The task is to evaluate the oral proficiency of a speaker as they read off a transcript. This paper expands upon the speech evaluation system setup that follows the general architecture in \cite{chao2022,wong2023gpdatauncertainty}. The baseline architecture is shown in \mbox{Fig.} \ref{fig:mean_pool}. This computes per-phoneme features (\mbox{i.e.} one feature vector for each phoneme in the sentence) of the Goodness-Of-Pronunciation (GOP), Log-Phoneme Posterior (LPP), Log-Posterior Ratio (LPR), SR, ND, and Phoneme Embedding (PE). Per-frame features of pitch and embeddings extracted from a weighted sum across layers of an SSL model are also computed. The SSL model here is wav2vec 2.0 XLSR-53 \cite{conneau2021}, which is pre-trained on multiple languages. The per-frame features are then converted to per-phoneme representations by mean-pooling over the frames that are aligned with each phoneme. These are then concatenated with the other per-phoneme features, and used as inputs to the speech evaluation model. The speech evaluation model first linearly projects the input to a reduced 24 dimensions, followed by 3 layers of 8-headed transformer encoder layers. Single-headed sentence pooling then converts the per-phoneme sequence to a sentence-level representation. There is then a ReLU hidden layer, followed by a softmax output that classifies between possible sentence-level oral proficiency scores.

Each of the feature types is now described in more detail. Let the time-normalised log-posterior probability of a phoneme be
\begin{equation}
\phi_i\left(\rho\right)=\frac{1}{t_\text{e}\left(\rho_i^\text{ref}\right)-t_\text{s}\left(\rho_i^\text{ref}\right)+1}\sum_{t=t_\text{s}\left(\rho_i^\text{ref}\right)}^{t_\text{e}\left(\rho_i^\text{ref}\right)}\log P\left(\rho_t=\rho\middle|\mathbf{O}\right),
\label{eq:phoneme_posterior}
\end{equation}
where $\rho$ is a phoneme, $\rho_i^\text{ref}$ is the $i$th correct canonical phoneme in the sentence that should have been read, $t_\text{s}$ and $t_\text{e}$ are the start and end frame indexes of the occurrence of that phoneme respectively, $t$ is a frame index, and $\mathbf{O}$ is the acoustic observation. When using a frame-synchronous ASR model, the times can be computed from a numerator lattice forced-alignment between the speech and the text that should have been read. In \cite{hu2015}, the per-frame phoneme posterior, $P\left(\rho_t\middle|\mathbf{O}\right)$, is computed as the softmax output of the neural network acoustic model, avoiding the computation expense of denominator lattice forward-backward operations. The duration normalisation encourages GOP features from phonemes with different durations to reside within similar dynamic ranges. The GOP is the time-normalised log-posterior probability of a canonical phoneme \cite{hu2015},
\begin{equation}
\text{GOP}_i=\phi_i\left(\rho_i^\text{ref}\right).
\label{eq:gop_exact}
\end{equation}
A commonly-used variant is to measure the margin between the log-posteriors of the canonical and most likely phonemes \cite{hu2015},
\begin{equation}
\text{GOP}_{\text{margin},i}=\phi_i\left(\rho_i^\text{ref}\right)-\max_{\rho}\phi_i\left(\rho\right).
\label{eq:gop_approx}
\end{equation}

LPP and LPR features are often also used for speech evaluation \cite{hu2015}. LPP is defined as a vector of the log-posteriors of all phonemes, such that
\begin{equation}
\text{LPP}_i=\left[\phi_i\left(\rho\right),\forall\rho\in\mathcal{P}\right],
\end{equation}
where $\mathcal{P}$ is the set of all phonemes. The LPR is the log-ratio between the posteriors of each phoneme and the canonical,
\begin{equation}
\text{LPR}_i=\left[\phi_i\left(\rho_i^\text{ref}\right)-\phi_i\left(\rho\right),\forall\rho\in\mathcal{P}\right].
\end{equation}
The GOP, LPP, and LPR provide information about confusability between phonemes that are pronounced, as perceived by the ASR model.

The SR is the rate at which the canonical phoneme is spoken \cite{huayunzhang2021},
\begin{equation}
\text{speaking rate}_i=\frac{1}{t_\text{e}\left(\rho_i^\text{ref}\right)-t_\text{s}\left(\rho_i^\text{ref}\right)}.
\end{equation}
The ND is the time taken to utter the canonical phoneme, normalised over all phonemes and utterances \cite{huayunzhang2021},
\begin{equation}
\text{normalised phoneme duration}_i=\frac{t_\text{e}\left(\rho_i^\text{ref}\right)-t_\text{s}\left(\rho_i^\text{ref}\right)-\mu}{\sigma},
\end{equation}
where $\mu$ and $\sigma$ are the mean and standard deviation of the un-normalised phoneme durations, $t_\text{e}-t_\text{s}$, respectively. These features are indicative of the speaker's fluency.

PE may represent semantic information, analogously to word embeddings \cite{mikolov2013}. These were extracted from a 32-node recurrent neural network continuous skip-gram language model \cite{mikolov2013}, which was trained on the non-silence canonical phoneme sequences of the speech evaluation datasets. Pitch was computed using the approach in \cite{ghahremani2014}, and may inform about the speaker's intonation.

\section{Computing features from Whisper}
\label{sec:whisper_features}

Open-source multi-lingual models allow for easy development of speech processing on a variety of languages, many of which that are supported are considered as low-resourced. For speech evaluation, this may simplify the expansion of applications to new languages. One such model is Whisper, which is trained in a weakly-supervised manner toward a variety of tasks, one of which is ASR. Whisper is designed with an attention encoder-decoder architecture \cite{chan2016}, which is frame-asynchronous and thus does not directly yield time alignments. Whisper also outputs formatted byte-pair encoding tokens, from which the computation of phoneme posteriors is not trivial. The formatted output may also not adequately express disfluencies, which may be useful information for speech evaluation. Work in \cite{ma2023} considers improving the disfluency expression of Whisper using prompt tuning, and is orthogonal to this paper.

Consider the text formatting and lack of expression of phonemes. The weakly-supervised model can be used to compute ASR hypotheses, expressed as an N-best list of word sequences. This paper proposes to use rule-based text normalisation to standardise casing, remove punctuations, and express numbers in spoken form. A lexicon is used to convert words into phoneme sequences. Although Whisper can be prompted to transcribe in a specified target language, words that do not belong to that language may still be found within the decoded results. It may be reasonable to assume that these foreign words are phonetically related to the input speech, and may therefore still be useful when computing phoneme confusions. Transliteration or romanisation can be used for non-Latin and Latin script target languages respectively, to convert these to similarly sounding character sequences in the script of the target language. A Grapheme-to-Phoneme (G2P) model can supplement the lexicon for out-of-vocabulary words. Multiple alternative pronunciations can be considered to improve generalisability, together with prior pronunciation probabilities computed from the G2P model.

This paper proposes to compute per-phoneme phoneme posteriors, $P\left(\rho_i\middle|\mathbf{O}\right)$, from a phoneme-level CN, instead of as \eqref{eq:phoneme_posterior}. A CN \cite{mangu2000,evermann2000a,xu2011} is a representation of competing decoded hypotheses, which assumes that the probability of the phoneme at each position along the sequence is conditionally independent of phonemes at other positions along the sequence. In this way, the probability of the phoneme sequence can be expressed as
\begin{equation}
P\left(\bm{\rho}\middle|\mathbf{O}\right)\approx\prod_iP\left(\rho_i\middle|\mathbf{O}\right).
\end{equation}
A CN can be approximated from an N-best list by using approaches of \cite{mangu2000,evermann2000a,xu2011}, with \cite{xu2011} being used in this paper. The CN can be interpreted to express a confidence in the model's hypothesis. A related approach in \cite{aggarwal2025} instead trains a Whisper-based model to directly predict confidence scores from a given hypothesis.

GOP-like features, referred to as CN-GOP, are computed from the log-posteriors of the phoneme-level CN,
\begin{equation}
\text{CN-GOP}_i=\log P\left(\rho_i=\rho_i^\text{ref}\middle|\mathbf{O}\right).
\label{eq:cn_gop_exact}
\end{equation}
Analogously to \eqref{eq:gop_approx}, a margin between the log-posteriors of the canonical and most likely phonemes can also be computed as
\begin{equation}
\text{CN-GOP}_{\text{margin},i}=\log P\!\left(\rho_i\!=\!\rho_i^\text{ref}\middle|\mathbf{O}\right)-\max\limits_{\rho}\log P\!\left(\rho_i\!=\!\rho\middle|\mathbf{O}\right)\!.
\label{eq:cn_gop_approx}
\end{equation}
Duration normalisation is omitted, to accommodate for the absence of phoneme time alignments from a frame-asynchronous model. Analogous LPP and LPR features can also be computed from the CN.

In summary, the proposed steps to compute the phoneme-level CN from frame-asynchronous non-phonemic formatted hypotheses are as follows.
\begin{enumerate}
\item Beam-search ASR decoding with the weakly-supervised model is used to compute an N-best list of formatted hypotheses.
\item The N-best word sequences are then text normalised.
\item Transliteration or romanisation is applied to accommodate for hypothesised words that are written in a script that differs from the target language.
\item The N-best list is filtered to remove identical word sequences, retaining that with the largest posterior.
\item The word sequences are converted to phoneme sequences through a lexicon.
\item A phoneme-level CN is computed from the phoneme-level N-best list. This is done by first expressing the N-best list as a phoneme lattice, with branched paths for competing N-best hypotheses and also branched paths for the alternative pronunciations of each word. The branch scores can be computed from the N-best posteriors and the G2P pronunciation priors. The algorithm in \cite{xu2011} is then used to compute a CN from the lattice.
\end{enumerate}

For read speech, GOP, LPP, and LPR features are computed for each canonical phoneme. This is computed by first finding a minimum edit distance alignment between the CN and the canonical phoneme sequence of the transcript. Each confusion set is categorised as either an insertion, deletion, or matched with a canonical phoneme. If matched, then the phoneme posterior can be read off from the confusion set, and phonemes that are not expressed in the confusion set are assumed to have zero probability. For a deletion, it is assumed that all probability is allocated to silence, and thus the non-silence phonemes are assigned zero probability. Insertions can be ignored, as GOP, LPP, and LPR features only need to be computed for each canonical phoneme.

A limitation is that pronunciation scores in the lexicon are priors, without considering the observed audio. Furthermore, the pronunciation variants present in the lexicon may not adequately capture idiosyncrasies present in non-native speech.

An alternative to beam-search decoding in step 1 is to instead generate hypotheses by sampling from the posterior. Initial tests found that for speechocean762, a high temperature needs to be used to obtain a set of diverse hypotheses. On the other hand, beam-search consistently yielded diverse hypotheses without needing to apply a temperature. Applying a temperature may impact the distribution expressed in the set of hypotheses with their sentence-level posteriors, which may then affect the GOP, LPP, and LPR features that are computed from these. This paper therefore favours using beam-search decoding over sampling generation.

The model's frame-asynchronous nature hinders the direct computation of the SR and ND. However, it is possible to estimate token-level time alignments through a dynamic time warping on the cross-attention weights. If the tokens are sub-word units, then word-level SR and ND features can be computed from these, and used instead of phoneme-level features.

\section{Alignment-free cross-attention}
\label{sec:alignment_free_model}

The speech evaluation models in \cite{chao2022,wong2023gpdatauncertainty} have inputs with sequence lengths corresponding to the number of canonical phonemes in the utterance. Per-frame features extracted from pre-trained SSL models and pitch can supplement the per-phoneme features. In \cite{chao2022,wong2023gpdatauncertainty}, this is achieved by mean pooling the per-frame features over the frames that are aligned to each canonical phoneme. An example of a model with this feature combination is shown in \mbox{Fig.} \ref{fig:mean_pool}.

An absence of a phoneme time alignment hinders the joint usage of phoneme-level and frame-level features. This paper proposes that a cross-attention, such as a transformer decoder \cite{vaswani2017}, can allow the joint usage of phoneme-level and frame-level features, without needing to supply a time alignment. An example of this is illustrated in \mbox{Fig.} \ref{fig:cross_attention}. The phoneme-level input sequence is used as the query to the cross-attention, thereby determining the output sequence length, while the frame-level input sequence is used as both the key and value. This may allow the model to learn how to associate between phonemes and frames, thereby learning to align. However, this approach allows each phoneme position to attend over the entire frame sequence. Learning from the limited training data about an adequate association of which frames belong to which canonical phoneme may be difficult. Word time alignments can be used to restrict each phoneme position to only attend to frames that are aligned to the word that the phoneme belongs to. Although such a constraint is less restrictive than using a phoneme time alignment, it is easier to obtain a word rather than a phoneme time alignment from Whisper.

\section{Experiments}

\subsection{Setup}

Experiments were performed on two language datasets, speechocean762 \cite{junbozhang2021} in English and an internal Tamil dataset that is also used in \cite{huayunzhang2021,wong2023nnmulti}. Speechocean762 comprises 2500 sentences of English read speech from 125 native Mandarin speakers in each of the train and test sets. The data is annotated with various types of proficiency scores at the phoneme, word, and sentence levels. The experiments in this paper only consider pronunciation accuracy, which is annotated by 5 human raters per sentence, along a scale from 0 to 10 for the sentence and word levels, and from 0 to 2 for the phoneme level. A consensus reference score was obtained by averaging the scores of the 5 raters, which differs from \cite{junbozhang2021} that instead uses the median score. The Tamil data comprises 2118 and 2093 sentences in the train and test sets respectively. These were read speech from 100 speakers from Singapore, between 9 to 16 years old. The sentence-level pronunciation accuracy was annotated by 4 raters along a scale from 1 to 5. A consensus reference score was again computed as the average of the multiple rater scores.

A baseline approach for speech evaluation is to compute GOP, LPP, LPR, SR, and ND from a frame-synchronous hybrid neural network-Hidden Markov model \cite{bourlard1994} ASR model. For speechocean762, this was trained following the setup in \cite{junbozhang2021}. The acoustic model used a Time Delay Neural Network (TDNN) \cite{waibel1989}, and was trained on the Librispeech 960 hours data \cite{panayotov2015}, following the standard Kaldi \cite{kaldi} recipe, up to the cross-entropy stage. For Tamil, a lattice-free \cite{povey2016} hybrid model was trained on 220 hours of internal read closed-talking speech from 710 adults, previously used in \cite{huayunzhang2021,wong2023nnmulti}. The acoustic model interleaved TDNN and long short-term memory layers \cite{peddinti2018}. Forced alignment of the speech evaluation data using these ASR models was used to compute the GOP, LPP, LPR, SR, and ND. SSL embeddings, PE, and pitch features were also computed, as described in section \ref{sec:features}.

The baseline speech evaluation architecture is illustrated in \mbox{Fig.} \ref{fig:mean_pool} and is referred to as the phoneme pooling model. The proposed alignment-free architecture shown in \mbox{Fig.} \ref{fig:cross_attention} and described in section \ref{sec:alignment_free_model} is referred to as the cross-attention model. The speech evaluation model was trained toward the cross-entropy criterion, with early stopping measured on a 10\% held-out validation set. During inference, the score was hypothesised as the mean of the discrete output posterior. Following \cite{junbozhang2021}, both the hypothesis and reference scores were first rounded to the closest integers before computing the Pearson's Correlation Coefficient (PCC) and Mean Squared Error (MSE) evaluation metrics.

\subsection{Features from Whisper}

The first experiment investigates the use of multiple alternative pronunciations and pronunciation priors when computing the GOP, LPP, and LPR from Whisper, each on its own without combining with other features. The speechocean762 test set was used. Whisper-large-v3 was decoded with a beam size of 200, using English prompted as the target language. Non-Latin script words were romanised using Unidecode. The CMU lexicon was used to map words to phoneme sequences, which contains up to 6 pronunciation variants for each word. A G2P model from \cite{novak2016} was used to supplement 6 pronunciation variants for each word that was not contained within the CMU lexicon, and also to compute pronunciation priors. This initial investigation of the stand-alone quality of the features followed the baseline setup presented in \cite{junbozhang2021},  where per-phoneme Support Vector Regressors (SVR) \cite{drucker1996} were trained on each of the feature types to compute the phoneme-level pronunciation accuracy.

\begin{table}[t]
\centering
\caption{Phoneme pronunciation accuracy of per-phoneme SVRs trained on GOP, LPP, and LPR from Whisper, using multiple pronunciations and priors}
\label{tab:num_pronunciations}
\tabcolsep=3.6pt
\begin{tabular}{cc|ccc}
\hline
Number of&Pronunciation&\multicolumn{3}{c}{PCC$\uparrow$ / MSE$\downarrow$ for Whisper feature}\\
pronunciations&prior&CN-GOP&CN-GOP\textsubscript{margin}&LPP \& LPR\\
\hline\hline
1&no&0.24 / 0.95&0.28 / \textbf{0.99}&0.42 / 0.16\\
up to 6&no&\textbf{0.29} / \textbf{0.88}&\textbf{0.41} / \textbf{0.99}&\textbf{0.48} / 0.15\\
up to 6&yes&0.27 / 0.92&0.34 / 1.00&\textbf{0.48} / \textbf{0.14}\\
\hline
\end{tabular}
\end{table}

The results in Table \ref{tab:num_pronunciations} show that using multiple pronunciations in the lexicon improves performance, while using pronunciation priors may not yield further gains. When pronunciation priors were not used, all pronunciation variants were treated as equally likely. The lack of benefit of pronunciation priors may be because the priors computed by the G2P model are independent of the acoustic observation. Subsequent experiments used multiple pronunciations without priors.

\begin{table}[h]
\centering
\caption{Compare features from hybrid model and Whisper with different beam sizes, to compute phoneme pronunciation accuracy, using per-phoneme SVRs}
\label{tab:beam_size}
\begin{tabular}{l|c|ccc}
\hline
Features&N-best&\multicolumn{3}{c}{PCC / MSE for feature}\\
from&size&GOP&GOP\textsubscript{margin}&LPP \& LPR\\
\hline\hline
hybrid&-&0.28 / 0.60&0.25 / 0.69&0.47 / 0.16\\
\hline
\multirow{3}{*}{Whisper}&10&0.31 / 0.91&0.37 / 0.99&0.49 / 0.17\\
&100&0.30 / 0.90&0.41 / 1.00&0.47 / 0.15\\
&200&0.29 / 0.88&0.41 / 0.99&0.48 / 0.15\\
\hline
\end{tabular}
\end{table}

The next experiment compares the Whisper-generated features using different beam sizes, against features computed from a hybrid model. Per-phoneme SVRs were again used to model phoneme pronunciation accuracy. The results in Table \ref{tab:beam_size} suggest that the feature quality may be fairly independent of the decoding beam size, with the only substantial improvement being in the PCC for CN-GOP\textsubscript{margin} when increasing the beam from 10 to 100. It therefore seems reasonable to use a smaller decoding beam to reduce computational cost. However, the remaining experiments used a beam size of 200, as the computation was still feasible with the dataset sizes that were used. The combination of LPP and LPR features computed from Whisper performs comparably to those computed from a hybrid model. For GOP\textsubscript{margin}, features extracted from Whisper have better PCC but worse MSE. The degraded MSE may be due to an interaction between the sparsity of the CN posteriors and the per-phoneme modelling of the SVRs.

\begin{table}[t]
\centering
\caption{Sentence-level pronunciation accuracy of phoneme pooling model on GOP, LPP, and LPR features from Whisper}
\label{tab:gop_whisper}
\begin{tabular}{llc|cc}
\hline
Features from&GOP type&Use LPP \& LPR&PCC&MSE\\
\hline\hline
-&-&no&0.709&1.138\\
hybrid&GOP\textsubscript{margin}&yes&0.753&1.051\\
Whisper&CN-GOP\textsubscript{margin}&yes&0.748&1.042\\
Whisper&CN-GOP&yes&0.745&1.026\\
\hline
\end{tabular}
\end{table}

A phoneme pooling model was then trained using the phoneme-level GOP\textsubscript{margin}, LPP, LPR, SR, ND, and PE features, together with the frame-level SSL and pitch features. This computed the sentence-level pronunciation accuracy. Table \ref{tab:gop_whisper} compares omitting the GOP, LPP, and LPR features (top row), as well as replacing these with features computed from Whisper. Omitting these three features degrades performance, showing their complementarity to the other features. Features computed from Whisper perform comparably to those computed from a hybrid model.

\begin{table}[h]
\centering
\caption{Inclusion of SR and ND features from Whisper}
\label{tab:tempo_whisper}
\begin{tabular}{l|cc}
\hline
Use SR and ND&PCC&MSE\\
\hline\hline
no&0.729&1.125\\
hybrid phoneme level&0.753&1.051\\
Whisper word level&0.743&1.031\\
\hline
\end{tabular}
\end{table}

Table \ref{tab:tempo_whisper} then investigates using word-level SR and ND features computed from Whisper alignments, using the same phoneme pooling model. Each canonical phoneme position was assigned the SR and ND of the word that the phoneme belonged to. The GOP\textsubscript{margin}, LPP, and LPR features here were computed from a hybrid model. The results again show that omitting SR and ND features degrades performance. Using word-level SR and ND computed from a Whisper alignment performs comparably against using phoneme-level SR and ND from a hybrid model.

\subsection{Alignment-free cross-attention model}
\label{sec:exp_cross_attention}

The models used thus far have combined the per-phoneme GOP\textsubscript{margin}, LPP, LPR, SR, ND, and PE features together with the per-frame SSL and pitch features, by averaging the per-frame features over the frames that are aligned with each canonical phoneme. A phoneme time alignment is not easily available from a frame-asynchronous model that does not explicitly express phonemes, like Whisper. Section \ref{sec:alignment_free_model} proposes a cross-attention approach, that allows for combination between these features without phoneme time alignments. This experiment investigates using this model architecture, shown in \mbox{Fig.} \ref{fig:cross_attention}. The per-phoneme and per-frame features were separately projected to 24 dimensions. An 8-headed 24-dimensional transformer decoder layer was used to combine these feature projections. The output of this was then parsed through 2 transformer encoder layers, followed by sentence-level attention pooling, ReLU feed-forward, and a softmax output.

\begin{table}[t]
\centering
\caption{Combine per-phoneme and per-frame features either by averaging over the frames for each phoneme or by using cross-attention}
\label{tab:alignment_free_model}
\tabcolsep=5.6pt
\begin{tabular}{l|l|cc}
\hline
Features from&Combine features by&PCC&MSE\\
\hline\hline
\multirow{2}{*}{hybrid}&phoneme pooling&0.753&1.051\\
&cross-attention&0.718&1.095\\
\hline
\multirow{2}{*}{Whisper}&cross-attention&0.735&1.051\\
&+ restrict attention to word&0.747&1.027\\
\hline
\end{tabular}
\end{table}

The results shown in Table \ref{tab:alignment_free_model} start from a baseline configuration of using features from a hybrid model and used the phoneme pooling architecture. The features were then kept the same, but the model architecture was replaced with the cross-attention approach. This yields a performance degradation, suggesting the benefit that a phoneme time alignment can bring when associating between the per-frame and per-phoneme features. While still using the cross-attention architecture, the hybrid model features were replaced with the CN-GOP\textsubscript{margin}, LPP, LPR, and word SR and ND features from Whisper. This configuration does not need a hybrid model. This performs similarly to the baseline for MSE, but still has a degraded PCC. The cross-attention allows each phoneme position to attend over all frames. Section \ref{sec:alignment_free_model} proposes that a word time alignment can be used to restrict the cross-attention, such that each phoneme only attends to frames that are aligned to the same word as that phoneme. The results in the bottom row of Table \ref{tab:alignment_free_model} suggest that such an attention constraint may be useful, and brings the performance back to being more comparable with the baseline. Thus, speech evaluation can be done without ASR training data.

\subsection{Word-level posteriors}

Section \ref{sec:whisper_features} describes how phoneme-level posteriors can be computed from a CN representation of ASR hypotheses. These can then be used to compute GOP, LPP, and LPR features for speech evaluation. Analogous posteriors can also be expressed at the word level, and be used to compute word-level GOP features. These may be complementary to the phoneme-level features. This experiment investigates the usefulness of these word-level GOP features.

The following steps were used to compute the features. The 200-best list from Whisper beam search was text normalised, romanised, and filtered for repetitions. A word-level CN was then computed from this N-best list. The CN was aligned with the reference transcription. For a confusion set that was matched with a reference word, the word posteriors could be read off the confusion set, with zero probability assigned to words that were not in the confusion set. For deletions, all non-silence word probabilities were assumed to be zero. Insertions were again ignored. Analogous word-level GOP-like features, referred to as CN-WGOP and CN-WGOP\textsubscript{margin} were computed from the word posteriors. As a sanity check, the PCC between CN-WGOP and the reference word-level pronunciation accuracy on the speechocean762 test set is 0.477, suggesting a reasonable correlation with the human-labelled word accuracy.

\begin{table}[h]
\centering
\caption{Inclusion of word GOP features in cross-attention model}
\label{tab:word_gop}
\begin{tabular}{l|cc}
\hline
Use word-level GOP&PCC&MSE\\
\hline\hline
no&\textbf{0.747}&\textbf{1.027}\\
CN-WGOP\textsubscript{margin}&0.745&1.040\\
CN-WGOP&0.737&1.077\\
\hline
\end{tabular}
\end{table}

A baseline setup used a cross-attention model with word time alignment constrained attention, using phoneme CN-GOP\textsubscript{margin}, LPP, LPR, and word SR and ND features from Whisper, together with PE, SSL, and pitch. The word-level GOP was then concatenated to the per-phoneme features, with each phoneme position in the sequence using the GOP of the word that the phoneme belonged to. Either CN-WGOP or CN-WGOP\textsubscript{margin} was used. The sentence-level pronunciation accuracy results in Table \ref{tab:word_gop} suggest that using CN-WGOP\textsubscript{margin} outperforms using CN-WGOP, but both do not yield performance improvements over the baseline. This suggests that word-level GOP may not be complementary to the existing phoneme-level and frame-level features for speech evaluation.

\subsection{Extend speech evaluation to low-resource Tamil language}

Using a multi-lingual open-source weakly-supervised model to compute features and time alignments for speech evaluation alleviates the need for paired speech-text data, simplifies extending speech evaluation to new languages supported by the weakly-supervised model, and may allow speech evaluation applications to benefit from the latest open-source model releases. The experiments thus far have considered speech evaluation in English, which is a reasonably high-resourced language. It is more difficult to obtain large quantities of diverse paired speech-text data for Tamil, to train an ASR model. Internal Tamil ASR training data needed to be collected here, because of the sparsity of open-source alternatives. It is often difficult to share such internal data, due to privacy constraints. Thus, Tamil is considered a low-resource language, and the proposal here can be applied when no Tamil ASR data is available at all. Whisper supports Tamil ASR, and thus can be used out-of-the-box to generate features and time alignments for speech evaluation. This section demonstrates the application of the proposed approach to Tamil speech evaluation.

In the previous experiments on English, hypothesised words from Whisper that were not in Latin script were text-normalised and romanised. The Tamil N-best lists were text-normalised to remove punctuation and convert numbers into spoken form. Open-source transliteration tools for Tamil as the target script are not as readily available as romanisation tools for a target with Latin script. As such, a simpler approach was used for Tamil by omitting N-best hypotheses that contained non-Tamil characters. The Tamil words were converted into phoneme sequences using a lexicon obtained from Babel \cite{harper2011}. This expresses up to 4 pronunciation variants for each word. A G2P model trained on the Babel lexicon was used to generate 4 pronunciation variants for hypothesised Tamil words that were not contained in the Babel lexicon.

\begin{table}[t]
\centering
\caption{Apply Whisper features and cross-attention model to low-resource Tamil language}
\label{tab:tamil}
\tabcolsep=5.6pt
\begin{tabular}{l|l|cc}
\hline
Features from&Combine features by&PCC&MSE\\
\hline\hline
\multirow{2}{*}{hybrid}&phoneme pooling&0.547&0.637\\
&cross-attention&0.611&0.699\\
\hline
\multirow{2}{*}{Whisper}&cross-attention&0.611&0.573\\
&+ restrict attention to word&0.629&0.569\\
\hline
\end{tabular}
\end{table}

The experiment from section \ref{sec:exp_cross_attention} is applied to Tamil in Table \ref{tab:tamil}. This compares a baseline hybrid model features approach and the proposed approach of computing features and time alignments from Whisper, on Tamil. As with the previous English experiments, the baseline approach used GOP\textsubscript{margin}, LPP, LPR, SR, ND, and phoneme forced alignments from the hybrid model. These were combined with PE, pitch, and SSL features. The pitch and SSL features were averaged over the frames that were aligned with each canonical phoneme. The proposed approach used CN-GOP\textsubscript{margin}, LPP, LPR, word-level SR and ND, and word-level time alignments computed from Whisper. Together with PE, these were combined with pitch and SSL features using cross-attention. Word posteriors were not used.

The results show that using a cross-attention architecture with features from a hybrid model improves PCC but degrades MSE, compared to a phoneme pooling architecture that uses a hybrid forced alignment. It is thus difficult to conclude how changing the architecture impacts performance. However, changing the features to those from Whisper improves both PCC and MSE, compared to hybrid features with a phoneme pooling model. Restricting the transformer decoder attention according to the Whisper word time alignment further improves the performance. This suggests that on a low-resource language, like Tamil, speech evaluation can be performed without any need for paired speech-text training data, by using an easily available multi-lingual open-source ASR model. This approach not only has comparable performance, but may surpass the performance of the hybrid baseline. Perhaps, in a low-resource scenario, it may be beneficial to rely on a multi-lingual open-source model that was trained on a large quantity of diverse data, rather than attempting to source for limited target language paired speech-text data to train an ASR model.

\subsection{Confusion network decoding of Whisper}

\begin{table}[t]
\centering
\caption{Confusion network decoding of Whisper N-best list}
\label{tab:cn_decode}
\begin{tabular}{c|cc}
\hline
N-best size&1-best WER (\%)$\downarrow$&CN decode WER (\%)\\
\hline\hline
1 (greedy)&17.40&-\\
10&16.25&15.66\\
100&16.18&15.41\\
200&16.22&\textbf{15.36}\\
\hline
\end{tabular}
\end{table}

Throughout this paper, CNs computed from Whisper N-best lists are repeatedly used to generate features for speech evaluation. With the availability of these CNs, it seems convenient to assess whether CN decoding for ASR has any benefit for a large, weakly-supervised, attention encoder-decoder model like Whisper. Table \ref{tab:cn_decode} shows the Word Error Rate (WER) measured on the speechocean762 test set when either selecting the top hypothesis from the N-best list or when first computing a CN, and then selecting the top word from each confusion set. The results suggest that CN decoding consistently outperforms selecting the 1-best, or top, hypothesis. When selecting the top hypothesis, increasing the decoding beam size only yields improvements up to a beam size of 100. For CN decoding, increasing the beam size continues to yield improvements up to the largest experimented beam size of 200. This suggests that applying CN decoding to a large, weakly-supervised model like Whisper may still be beneficial.

\section{Conclusion}

This paper has considered how to alleviate the reliance of speech evaluation on expensive target language paired speech-text data to train a phonemic frame-synchronous ASR model, by relaxing the requirements to allow the use of more easily obtainable open-source multi-lingual frame-asynchronous models. Alternative GOP, SR, and ND features have been proposed without requiring phoneme time alignments, and instead using CNs. A cross-attention model architecture has also been proposed to combine between per-phoneme and per-frame features. The experiments show that these perform comparably to the hybrid baseline on English speechocean762, and even outperform the baseline on low-resource Tamil. Thus, speech evaluation can be done without expensive speech-text data.

\bibliographystyle{IEEEbib}
\bibliography{refs}

\end{document}